%% file: ms.tex
\documentclass[sigconf]{acmart}

\pdfoutput=1

\usepackage{booktabs} 

\setcopyright{none}

\renewcommand\footnotetextcopyrightpermission[1]{} 
\usepackage{balance}  
\usepackage{color}
\usepackage{amsmath}
\usepackage{tabularx}
\usepackage{multirow}
\newcolumntype{L}{>{\raggedright\arraybackslash}X} 
\usepackage{tikz} 
\usetikzlibrary{shapes,shapes.arrows,shapes.geometric,shapes.multipart,positioning,fit}
\usetikzlibrary{shapes.multipart}  
\usepackage{caption}
\usepackage{subcaption}
\usepackage{listings}
\usepackage{algorithm}
\usepackage{algpseudocode}

\newtheorem{theorem}{Theorem}

\newtheorem{definition}[theorem]{Definition}

\newcommand{\Figure}[1]{Figure~\ref{#1}}

\newcommand\Section[1]{Section~\ref{#1}}

\begin{document}
\title{TFLMS: Large Model Support in TensorFlow by Graph Rewriting}

\author{Tung D. Le}
\affiliation{
  \institution{IBM Research - Tokyo}
  \streetaddress{19-21, Nihonbashi Hakozaki-cho, Chuo-ku}
  \city{Tokyo}
  \country{Japan}
  \postcode{103-8510}
}
\email{tung@jp.ibm.com}

\author{Haruki Imai}
\affiliation{
  \institution{IBM Research - Tokyo}
  \streetaddress{19-21, Nihonbashi Hakozaki-cho, Chuo-ku}
  \city{Tokyo}
  \country{Japan}
  \postcode{103-8510}
}
\email{imaihal@jp.ibm.com}

\author{Yasushi Negishi}
\affiliation{
  \institution{IBM Research - Tokyo}
  \streetaddress{19-21, Nihonbashi Hakozaki-cho, Chuo-ku}
  \city{Tokyo}
  \country{Japan}
  \postcode{103-8510}
}
\email{negishi@jp.ibm.com}

\author{Kiyokuni Kawachiya}
\affiliation{
  \institution{IBM Research - Tokyo}
  \streetaddress{19-21, Nihonbashi Hakozaki-cho, Chuo-ku}
  \city{Tokyo}
  \country{Japan}
  \postcode{103-8510}
}
\email{kawatiya@jp.ibm.com}

\begin{abstract}
  \input{abstract}
\end{abstract}

%
%

\maketitle

\section{Introduction}
\label{sec:intro}
\input{introduction}

\section{Related work}
\label{sec:relw}
\input{related-work}

\section{Computational graphs}
\label{sec:comp-graph}
\input{comp-graph}

\section{Graph rewriting}
\label{sec:grprw}
\input{graph-rewriting}

\section{TFLMS module in TensorFlow}
\label{sec:tflms}
\input{tflms}

\section{Experiments}
\label{sec:exprm}
\input{experiments}

\section{Conclusion}
\label{sec:concl}
\input{conclusion}

\begin{acks}
\input{acks}
\end{acks}

\balance
\bibliographystyle{ACM-Reference-Format}
\bibliography{refs}

\end{document}

%% file: abstract.tex
While accelerators such as GPUs have limited memory, deep neural networks are becoming larger and will not fit with the memory limitation of accelerators for training.
We propose an approach to tackle this problem by rewriting the computational graph of a neural network, in which swap-out and swap-in operations are inserted to temporarily store intermediate results on CPU memory.
In particular, we first revise the concept of a computational graph by defining a concrete semantics for variables in a graph.
We then formally show how to derive swap-out and swap-in operations from an existing graph and present rules to optimize the graph.
To realize our approach, we developed a module in TensorFlow, named TFLMS.
TFLMS is published as a pull request in the TensorFlow repository for contributing to the TensorFlow community.
With TFLMS, we were able to train ResNet-50 and 3DUnet with $4.7x$ and $2x$ larger batch size, respectively.
In particular, we were able to train 3DUNet using images of size of $192^3$ for image segmentation, which, without TFLMS, had been done only by dividing the images to smaller images, which affects the accuracy.


%% file: introduction.tex
Deep neural networks together with deep learning are effective for solving complex signal-processing problems such as those in computer vision, speech recognition, and natural language processing.
However, training a neural network is time-consuming, often taking days to weeks.
The training is mainly based on matrix multiplications; therefore, it is often accelerated using accelerators such as GPUs.
In $2012$, GPUs were used for training a neural network for the first time.
It was a deep convolutional neural network of $16$ layers, called AlexNet~\cite{NIPS2012_4824}, achieving outstanding image classification results in the ILSVRC-2012 competition~\footnote{\url{http://www.image-net.org/challenges/LSVRC/2012/}} with a \mbox{top-5} test error rate of $15.3\%$.
Since then, GPUs have been popular for deep learning.

After the success of AlexNet in the ILSVRC-2012 competition, deep learning has evolved quickly for a broader spectrum of applications.
Neural networks are deeper (including more layers) and larger, e.g., ResNet-1001 consists of $1001$ layers and is much deeper than AlexNet~\cite{He:2016:ECCV}.
Thus, neural networks are sometimes too large to be fit with the memory limitation of GPUs for training.

From the hardware viewpoint, GPUs should be designed to have a larger physical memory, but increasing physical memory is expensive.
From the software viewpoint, there are three main approaches to solving this problem.
The first one is reducing memory consumption by reusing memory regions~\cite{Shirahata:MLSP:2016} for different computations, compressing a neural network~\cite{Choi:compress:2018} or using low precision~\cite{Faraone:precision:2017}, 
the second is re-computing some of the computations from checkpoints~\cite{2016arXiv160406174C},
the third is using an external memory such as CPU memory for temporarily storing intermediate results during training~\cite{Rhu:vDNN:2016, meng2017training}.

We pursued the third approach of using an external memory because it often helps with training a larger model compared to the other approaches and it can be generally applied to any neural networks.
Different from the previous studies involving swapping data from GPU memory to an external memory, and vice versa, in an ad-hoc manner, we propose an approach based on formal rules for graph rewriting, which is provable.
Our contributions in this paper are as follows:
\begin{itemize}
\item We revised the concept of a computational graph of a neural network.
  Our definition of a computational graph is inspired by that in TensorFlow~\cite{Abadi:MAPL:2017}, a popular framework for deep learning.
  Different from a computational graph in TensorFlow, variables in our computational graph are first-class citizens and consistent with the concept of operations in a computational graph.
\item We formally derived swap-out and swap-in operations from an existing graph, those used to exchange intermediate results between GPUs and CPUs.
  Derivation is based on some rules in program transformations with correctness guarantee, which helps us understand the nature of swapping operations.
\item We presented two strategies for finding control operations that are used to control when data are swapped in from an external memory to GPU memory, which helps improve performance.
\item To realize our approach, we developed a module in TensorFlow, called TFLMS.
  TFLMS is published as a pull request in the TensorFlow repository for contributing to the TensorFlow community.
  With TFLMS, we were able to train ResNet-50~\cite{He:2015:CORR} and 3DUnet~\cite{3dunet:2016} with a $4.7x$ and $2x$ larger batch size, respectively.
  In particular, we were able to train 3DUNet using images of size of $192^3$ for image segmentation, which, without TFLMS, had been done only by dividing the images to smaller images.
\end{itemize}

The rest of the paper is organized as follows.
In \Section{sec:relw}, we discuss related work.
In \Section{sec:comp-graph}, we discuss our proposed approach involving revising the concept of a computational graph and presenting the semantics of the graph.
In \Section{sec:grprw}, we discuss the rules to derive swap-out and swap-in operations and optimizations.
In \Section{sec:tflms}, we discuss our TFLMS module that implements our approach in TensorFlow.
In \Section{sec:exprm}, we present the experimental results.
\Section{sec:concl} summarizes the key points and discusses future work.


%% file: related-work.tex
The most intuitive method for training large models is using Unified Memory~\cite{um:gtc2017}, a single memory address space accessible from both CPUs and GPUs.
Enabling Unified Memory is simple, but its performance is very poor compared to custom methods that manually offload and prefetch data.
\citeauthor{Shirahata:MLSP:2016}~\cite{Shirahata:MLSP:2016} proposed a reduction approach of reusing, during the backward phase, the memory regions allocated for the forward phase.
\citeauthor{Rhu:vDNN:2016}~\cite{Rhu:vDNN:2016} proposed a different approach of managing runtime memory by virtualizing the memory usage of neural networks against both GPU and CPU memory.
During training, only the current layer is active and consumes GPU memory while the other layers' data are swapped out to the CPU memory. 
This approach performed better than using Unified Memory.
\citeauthor{meng2017training}~\cite{meng2017training} took the same approach as~\cite{Rhu:vDNN:2016} for TensorFlow by swapping tensors from GPU memory to CPU memory and vice versa.
However, the authors did not discuss how to derive swap-out and swap-in operations~\cite{meng2017training}.
Besides, we could not find their TensorFlow source code.
We borrowed \citeauthor{meng2017training}'s idea but formally defined transformation rules for graph rewriting so that the correctness of the transformed computational graph is provable.
Apart from using CPU memory as a temporary memory for computation, \citeauthor{2016arXiv160406174C}~\cite{2016arXiv160406174C} proposed an approach of gradient-checkpointing, in which checkpointing vertices in a computational graph are automatically defined using graph partition.
Parts of the graph in between checkpointing vertices are re-computed during the backward phase.
The forward phase is generally computed twice.
\citeauthor{Wang:2018:SDG:3178487.3178491}~\cite{Wang:2018:SDG:3178487.3178491} combined both swapping and recomputation in a single framework.



%% file: comp-graph.tex
\begin{figure}[!t]
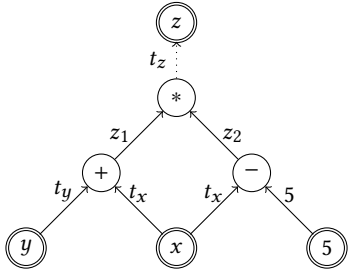

  \centering
  \tikz [scale=1,
  every node/.style={scale=1},
  ops/.style={draw, circle, minimum size=0.5cm, inner sep=0},
  placeholder/.style={draw, double, circle, minimum size=0.5cm, inner sep=0}] {
    \node[placeholder] (z) at (2, 3) {$z$};
    \node[ops] (mul) at (2, 2) {$*$};
    \node[ops] (plus1) at (1, 1) {$+$};
    \node[ops] (plus2) at (3, 1) {$-$};
    \node[placeholder] (x) at (2,0) {$x$};
    \node[placeholder] (y) at (0,0) {$y$};
    \node[placeholder] (c) at (4, 0) {$5$};
    \draw[->] (x) edge node[above] {$t_x$} (plus1);
    \draw[->] (y) edge node[above] {$t_y$} (plus1);
    \draw[->] (x) edge node[above] {$t_x$} (plus2);
    \draw[->] (c) edge node[above] {$5$} (plus2);
    \draw[->] (plus1) edge node[left] {$z_1$} (mul);
    \draw[->] (plus2) edge node[right] {$z_2$} (mul);
    \draw[->, dotted] (mul) edge node[left] {$t_z$} (z);
  }
  \caption{Computational graph for $z = (x + y) * (x - 5)$.
    Vertices are operations and edges are tensors.
    Continuous arrows represent "read" edges and dotted arrows represent "update" edges.
    Double circles are parameterized operations including variables and constants.}
  \label{fig:comp-graph}
\end{figure}

\begin{table*}[ht]
  \centering
  \caption{Notations}  
  \tikzstyle{every node}=[scale=1]
  \tikzstyle{ops}=[draw, circle, minimum size=0.5cm, inner sep=0]
  \tikzstyle{placeholder}=[draw, double, circle, minimum size=0.5cm, inner sep=0]
  \begin{tabularx}{\linewidth}{cclL}
    \toprule
    & Notation & Definition & Meaning \\
    \midrule
    \multirow{2}{*}[-10pt]{Vertices}
    &
    \tikz[baseline] {
    \node[ops] (op1) {$f$};
    \node[above left of=op1] (in1) {};
    \node[left of=op1] (in2) {};
    \node[below left of=op1] (in3) {};
    \node[right of=op1] (out1) {};
    \draw[->] (in1) -- node[above] {$t_1$} (op1);
    \draw[->] (in2) -- node[above] {$t_2$} (op1);
    \draw[->] (in3) -- node[above] {$t_3$} (op1);
    \draw[->] (op1) -- node[above] {$t_4$} (out1);
    }
             & $
               \begin{aligned}
                 \lambda (f) &= (f, False) \\
                 t_4 &= f([t_1, t_2, t_3])
               \end{aligned}$
               & Normal operation $f$, taking list of inputs $t_1,t_2, t_3$ and returning output $t_4$.  \\
    \cmidrule{2-4}
    &
    \tikz[baseline] {
    \node[placeholder] (op1) {$x$};
    \node[right of=op1] (op2) {};
    \draw[->] (op1) -- node[above] {$t$} (op2);
    }
             & $
               \begin{aligned}
                 [t]\lambda (x) &= (x, True) \\
                 t&=x
               \end{aligned}$
                          & Parameterized operation or variable $x$ . \\
    \midrule
    \multirow{3}{*}[-12pt]{Edges}
    &
    \tikz[baseline] {\node[ops] (op1) {$f$}; \node[ops, right of=op1] (op2) {$g$};
      \draw[->] (op1) -- node[above] {$t$} (op2)} & $\tau (f, g) = (t, "read")$ & $g$ reads $t$ as input. $t$ is output of $f$. This edge represents $g \circ f$.\\
    \cmidrule{2-4}
    &
    \tikz[baseline] {
    \node[ops] (op1) {$f$}; \node[placeholder, right of=op1] (op2) {$x$};
    \draw[->, dotted] (op1) -- node[above] {$t$} (op2)}
             & $
               \begin{aligned}[t]\tau (f, x) &= (t, "update") \\
                 x&=t
               \end{aligned}$
                          & $f$ produces output $t$ that is used to update variable $x$. $t$ and $x$ must have the same type. \\
    \cmidrule{2-4}
    &
    \tikz[baseline] {\node[ops] (op1) {$f$}; \node[ops, right of=op1] (op2) {$g$};
      \draw[->, dashed] (op1) -- node[above] {$$} (op2)} & $\tau (f, g) = (\_, "control")$ & $g$ cannot be executed unless $f$ finished.\\
    \bottomrule
  \end{tabularx}
  \label{tab:notations}
\end{table*}

\begin{figure*}[!ht]
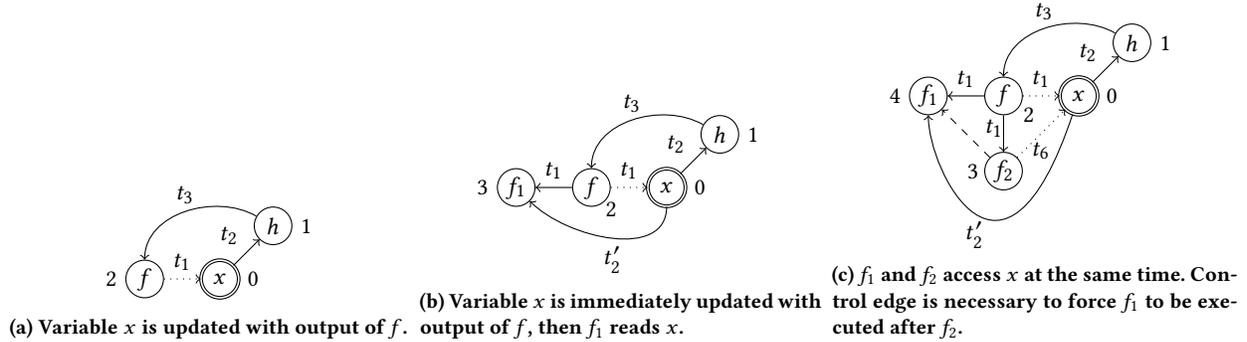

  \centering
  \setlength\tabcolsep{2pt}
  \begin{tabular}{ccc}
    \begin{subfigure}[b]{0.3\textwidth}
      \centering
      \tikzstyle{every node}=[scale=1]
      \tikzstyle{ops}=[draw, circle, minimum size=0.5cm, inner sep=0]
      \tikzstyle{placeholder}=[draw, double, circle, minimum size=0.5cm, inner sep=0]
      \tikzstyle{update}=[->, dotted]
      \tikzstyle{read}=[->]
      \tikz[baseline] {
        \node[ops, label={[name=lop1] left:2}] (op1) {$f$};
        \node[placeholder, right of=op1, label={[name=lop2] right:0}] (op2) {$x$};
        \node[ops, above right of=op2, label={[name=lop3] right:1}] (op3) {$h$};
        \draw[update] (op1) -- node[midway, auto] {$t_1$} (op2);
        \draw[read] (op2) -- node[midway, auto] {$t_2$} (op3);
        \draw[read] (op3) to [out=150, in=90] node[midway, above] {$t_3$} (op1);
      }
      \caption{Variable $x$ is updated with output of $f$.}
      \label{fig:eg1}
    \end{subfigure}
    &
      \begin{subfigure}[b]{0.3\textwidth}
        \centering
        \tikzstyle{every node}=[scale=1]
        \tikzstyle{ops}=[draw, circle, minimum size=0.5cm, inner sep=0]
        \tikzstyle{placeholder}=[draw, double, circle, minimum size=0.5cm, inner sep=0]
        \tikzstyle{update}=[->, dotted]
        \tikzstyle{read}=[->]
        \tikz[baseline] {
          \node[ops, label={[name=lop1, yshift=3pt, xshift=-3pt] below right:2}] (op1) {$f$};
          \node[placeholder, right of=op1, label={[name=lop2] right:0}] (op2) {$x$};
          \node[ops, above right of=op2, label={[name=lop3] right:1}] (op3) {$h$};
          \node[ops, left of=op1, label={[name=lop4] left:3}] (op4) {$f_1$};
          \draw[update] (op1) -- node[midway, auto] {$t_1$} (op2);
          \draw[read] (op2) -- node[midway, auto] {$t_2$} (op3);
          \draw[read] (op3) to [out=150, in=90] node[midway, above] {$t_3$} (op1);
          \draw[read] (op1) to node[midway, above] {$t_1$} (op4);
          \draw[read] (op2) to [out=270, in=-45] node[midway, below] {$t_2^{\prime}$} (op4);
        }
        \vspace*{-1mm}
        \caption{Variable $x$ is immediately updated with output of $f$, then $f_1$ reads $x$.}
        \label{fig:eg2}
      \end{subfigure}
    &
      \begin{subfigure}[b]{0.3\textwidth}
        \centering
        \tikzstyle{every node}=[scale=1]
        \tikzstyle{ops}=[draw, circle, minimum size=0.5cm, inner sep=0]
        \tikzstyle{placeholder}=[draw, double, circle, minimum size=0.5cm, inner sep=0]
        \tikzstyle{update}=[->, dotted]
        \tikzstyle{control}=[->, dashed]  
        \tikzstyle{read}=[->]
        \tikz[baseline] {
          \node[ops, label={[name=lop1, shift={(-1pt,4pt)}] below right:2}] (op1) {$f$};
          \node[placeholder, right of=op1, label={[name=lop2] right:0}] (op2) {$x$};
          \node[ops, above right of=op2, label={[name=lop3] right:1}] (op3) {$h$};
          \node[ops, left of=op1, label={[name=lop4] left:4}] (op4) {$f_1$};
          \node[ops, below of=op1, label={[name=lop5, xshift=1pt] left:3}] (op5) {$f_2$};
          \draw[update] (op1) -- node[midway, auto] {$t_1$} (op2);
          \draw[update] (op5) -- node[midway, below] {$t_6$} (op2);
          \draw[read] (op2) -- node[midway, auto] {$t_2$} (op3);
          \draw[read] (op3) to [out=150, in=90] node[midway, above] {$t_3$} (op1);
          \draw[read] (op1) to node[midway, above] {$t_1$} (op4);
          \draw[read] (op1) to node[midway, left, xshift=3pt, yshift=2pt] {$t_1$} (op5);
          \draw[read] (op2) .. controls +(-1,-3) and +(0,-1) .. node[midway, below] {$t_2^{\prime}$} (op4);
          \draw[control] (op5) -- node[midway, below] {$$} (op4);
        }
        \vspace*{-10mm}
        \caption{$f_1$ and $f_2$ access $x$ at the same time. Control edge is necessary to force $f_1$ to be executed after $f_2$.}
        \label{fig:eg3}
      \end{subfigure}
  \end{tabular}
  \caption{Examples about graphs regarding variables.
    Integers above or below vertex are order of that vertex in topological ordering.}
  \label{fig:variables}
\end{figure*}

A computational graph is a core concept in TensorFlow.
Neural networks defined by users are represented by a computational graph of operations.
TensorFlow then executes optimizations over the graph before invoking operations in the graph.
In this section, we revise the concept of a computational graph in TensorFlow~\cite{Abadi:MAPL:2017} to make its semantics more consistent.

\subsection{Definition}
\begin{definition}{(Computational graph)}
  Let $G = (V, E, \lambda, \tau)$ be a vertex and edge-labeled directed graph,
  where $V$ is the set of vertices in $G$, $E \subseteq V \times V$ is the set of edges in G,
  $\lambda : V \rightarrow (O, Bool)$ is a function mapping each vertex to a tuple of an operation $o \in O$ and a Boolean value indicating whether the operation is parameterized or not, and
  $\tau : E \rightarrow (T, {ACT})$ is a function mapping each edge to a tuple of a value of data type $T$ and an action in ${ACT}$ where ${ACT} = \{"read", "update", "control"\}$.
\end{definition}

Computational graphs are a way to express mathematical expressions in which each vertex is an operation with inputs of incoming edges and outputs of outgoing edges.
In deep learning, computational graphs are used to express computations in neural networks that consist of operations whose input and output are often multi-dimensional arrays.
Multi-dimensional arrays are often called \emph{tensors}.
Tensors that are used to store the internal states of a neural network, e.g., learning weights and bias in hidden layers in a neural network, are updated regularly.
Hence, we classify operations into \emph{normal operations} and \emph{parameterized operations} where parameterized operations have internal states that can be updated.
A \emph{variable} is a special parameterized operation that is to update its internal variable using the identity operation\footnote{Identity function accepts a value and returns the same value}.
A \emph{constant} is a special case of a variable where its value is set once and is never updated.
Each edge has a value indicating an action related to the tensor on the edge.
There are three actions: ``read'', ``update'', and ``control''.
Considering an edge $(u,v)$ from an operation $u$ to an operation $v$,
actions ``read'' and ``update'' mean $v$ reads and updates the tensor, respectively; and
action ``control'' means $u$ triggers the execution of $v$, and $u$ is called a \emph{control dependency operation}.

\Figure{fig:comp-graph} shows a computational graph for an expression $z = (x + y) * (x - 5)$.
In this example, there are three variables $x, y$, and $z$.
An outgoing edge emanating from a variable means reading the variable value, and an incoming edge to a variable means updating a tensor to the variable (denoted with a dotted arrow).

\subsection{Notations and semantics}
Table~\ref{tab:notations} lists the notations to represent different vertices and edges in a graph.
Function composition is denoted as ``$\circ$'', and, from its definition, we have $(f_2 \circ f_1) x = f_2(f_1(x))$.
Function ``$.\_i$'' is to take the $i$-th element in a tuple, e.g., $(a,b).\_2$ returns $b$.

An operation in a computational graph is generally triggered to execute when all of its incoming edges
have data.
The operation generates data on its outgoing edges then other operations are repeatedly triggered in the same manner.
This procedure ends when all of the reachable operations are executed and all of the reachable edges are filled with data.
In other words, each of the reachable operations, except variables, is executed once.

However, there is no way to trigger the execution of a graph.
At the beginning of computation, there is no way to set a value for an edge.
Furthermore, computational graphs are acyclic graphs, and there are some operations with no incoming edges.
These operations cannot be triggered.
This problem is resolved using variables.

Variables in a computational graph are used to store learnable parameters, input and output data, and are used to trigger computation of the graph.
Variables are special and make a computational graph for deep learning different from a general dependency graph.
Because a variable has an internal state, defining its semantics is non-trivial in the context of the graph.
At the beginning, variables are initialized with values input by users or random values generated by a distribution.
During training, they are updated by a learning optimizer.
This leads to a variable being visited more than once, and may introduce cycles if its semantics is ambiguous.
The remainder of this section introduces a clear semantics for variables.

To describe the semantics of a computational graph containing variables, we first define a topological ordering over a computational graph.
\begin{definition}{(Topological ordering)}
  Given a computational graph $G = (V, E, \lambda, \tau)$, let $N$ be the number of vertices in the graph, topological ordering is a mapping of vertices to an integer, $\gamma : V \rightarrow \{0, 1,\dots,N-1\}$, satisfying
  \begin{itemize}
  \item $\gamma(v) = 0, \forall v \in V \land \lambda(v).\_2 = \mathtt{True}$, and
  \item $\gamma(u) < \gamma(v), \forall (u,v) \in E \land \lambda(v).\_2 = \mathtt{False}$.
  \end{itemize}
\end{definition}

In general, a topological ordering represents the order of execution of operations in a graph.
Given two operations $u$ and $v$, if $\gamma(u) < \gamma(v)$, $u$ is executed before $v$.
If $\gamma(u) = \gamma(v)$, $u$ and $v$ are executed in parallel.
In this paper, variables always have order of $0$, which means variables will be executed first, and incoming edges (``update'' edges) to them do not change their order.
Later executions of a variable depend on its incoming operations, and are independent of the variable's order.
These executions alone do not trigger the variable's outgoing operations.

Let us consider the example graph in Figure~\ref{fig:eg1}.
The graph has the following execution ordering: ``$x \rightarrow h \rightarrow f \rightarrow x$''.
First, variable $x$ is initialized by users then it triggers operation $h$.
Then, $h$ is executed and triggers operation $f$.
Finally, $x$ is updated with the output of $f$, and the computation finishes.
Operation $h$ depends on $x$ only, and $x$ itself can not trigger $h$ again.

The example graph in Figure~\ref{fig:eg2} may have two possible execution orderings: ``$x \rightarrow h \rightarrow f \rightarrow x \rightarrow f_1$'', or ``$x \rightarrow h \rightarrow f \rightarrow f_1 \rightarrow x$''.
Operation $f_1$ is triggered based on the availability of tensors $t_1$ and $t_2^{\prime}$.
It is easy to see that $f_1$ must be executed after $f$ and after $x$.
However, $x$ is executed multiple times.
It is important to know which output of $x$ is used as input to $f_1$.

To avoid ambiguity, we present the following convention regarding variables:
\begin{itemize}
\item An operation is always using the latest value of a variable.
\item Variables always have the highest priority of execution among operations consuming the same tensor.
\end{itemize}
This convention helps us ensure that $f_1$ is executed after updating $x$ with the output from $f$.

The execution order of an operation not only depends on data availability on incoming edges but also control dependency edges.
``Control'' edges do not have data.
In other words, they are not inputs for the operation.
``Control'' edges are used to control the execution order of an operation.
Adding a ``control'' edge into a graph will alter the topological ordering of the graph.
If $(u,v)$ is a ``control'' edge, $v$ must be executed after $u$, and $\gamma(u) < \gamma(v)$.
By this definition, there is no control edge to a variable.

The example graph in Figure~\ref{fig:eg3} has a new operation, $f_2$, that consumes the output of $f$, executes computation and, updates variable $x$. 
Without the control edge from $f_2$ to $f_1$, after $f$ is executed, $f_1$ and $f_2$ can be executed in parallel because they do not depend on each other.
Because they both access variable $x$, i.e, $f_1$ reads $x$ and $f_2$ writes to $x$, a control edge is necessary to ensure that they access $x$ in order.
The ``control'' edge from $f_2$ to $f_1$ states that $f_1$ will be executed after finishing $f_2$ and updating $x$.

\subsection{Training using back-propagation}
Training a neural network involves minimizing an objective function $f$ measuring the distance between a ground truth value and predicted value.
The objective function is a composition of multiple functions with learnable parameters,
and the gradient descent algorithm is often used to minimize the function.
Optimization is an iterative procedure updating learnable parameters so that the objective function is minimized, in which each training iteration consists of three phases:
\emph{forward phase} to compute the objective function,
\emph{backward phase} to compute gradients of the objective function with respect to learnable parameters,
and \emph{update phase} to update learnable parameters using the gradients.
Backward phase is done via back-propagation for efficiency, starting from the objective function and propagating back gradients through the functions.
At the beginning of an iteration, tensors are cleaned up except variables for learnable parameters.
Variables for input tensors are fed with new data and trigger the iteration.
Because a training dataset is often very big, each iteration takes only a subset (\emph{batch}) of examples extracted from the training dataset as its input tensor.
The number of examples in a batch (or \emph{batch size}) will affect the size of the input tensor and also other tensors in the computational graph.
In general, increasing batch size will make a model larger.

\begin{figure}[!t]
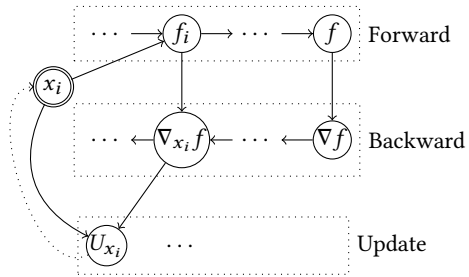

  \centering
  \tikzstyle{every node}=[scale=1]
  \tikzstyle{ops}=[draw, circle, minimum size=0.5cm, inner sep=0]
  \tikzstyle{placeholder}=[draw, double, circle, minimum size=0.5cm, inner sep=0]
  \tikzstyle{update}=[->, dotted]
  \tikzstyle{control}=[->, dashed]  
  \tikzstyle{read}=[->]

  \tikz[baseline] {
    \node (in) {$\dots$};
    \node[ops, right of=in] (fi) {$f_i$};
    \node[placeholder, below left of=in] (x) {$x_i$};
    \node[right of=fi] (fio) {$\dots$};
    \node[ops, right of=fio] (loss) {$f$};

    \node[below right of=x] (bin) {$\dots$};
    \node[ops, right of=bin] (gfi) {$\nabla_{x_i} f$};
    \node[right of=gfi] (gfii) {$\dots$};
    \node[ops, right of=gfii] (gloss) {$\nabla f$};

    \node[ops, below=of bin] (uf) {$U_{x_i}$};
    \node[below=of gloss] (n1) {};
    \node[right of=uf] (n2) {$\dots$};

    \draw[read] (in) -- node[midway, auto] {} (fi);
    \draw[read] (x) -- node[midway, auto] {} (fi);
    \draw[read] (fi) -- node[midway, auto] {} (fio);
    \draw[read] (fio) -- node[midway, auto] {} (loss);
    \draw[read] (loss) -- node[midway, auto] {} (gloss);
    \draw[read] (gloss) -- node[midway, auto] {} (gfii);
    \draw[read] (fi) -- node[midway, auto] {} (gfi);
    \draw[read] (gfii) -- node[midway, auto] {} (gfi);
    \draw[read] (gfi) -- node[midway, auto] {} (bin);
    \draw[read] (gfi) -- node[midway, auto] {} (uf);
    \draw[read] (x) .. controls +(-0.5,-1) and +(-1,0.5) .. node[midway, auto] {} (uf);
    \draw[update] (uf) .. controls +(-1,-0.5) and +(-1,0) .. node[midway, auto] {} (x);

    \node[draw,dotted,fit=(in) (fi) (fio) (loss), label=right:Forward] {};
    \node[draw,dotted,fit=(bin) (gfi) (gfii) (gloss), label=right:Backward] {};
    \node[draw,dotted,fit=(uf) (n1) (n2), label=right:Update] {};
  }
  \caption{How variable is used and updated in training.}
  \label{fig:training}
\end{figure}

Figure~\ref{fig:training} shows how learnable parameters (represented by variables) are updated during training.
In the forward phase, variable $x_i$ is an input to function $f_i$, outputs from $f_i$ are used in the later function, finally a loss value is produced by objective function $f$.
In the backward phase, we compute gradients of $f$ with respect to learnable parameters.
Function $\nabla_{x_i} f$ computes the gradient of $f$ with respect to $x_i$, which requires $f_i$'s output as one of its inputs.
Finally, $x_i$ is updated by a function $U_{x_i}$ during the update phase.

\subsection{Device placement}
In TensorFlow, each operation in the computational graph is placed on a device such as a GPU, CPU, FPGA.
Communication between two devices automatically occurs if an operation on one device consumes a tensor produced by another operation on the other device.
In fact, TensorFlow adds a pair of two operations, ``send'' and ``receive'', to the graph for exchanging a tensor.
In this paper, we do not show these communication operations when drawing graphs.

\subsection{Garbage collection}
If a tensor is no longer used in TensorFlow, it is released by TensorFlow garbage collection.
Every tensor is assigned a reference count, which is the number of operations.
Each time a tensor is consumed by an operation, its reference count is decreased by one.
If the reference count reaches zero, the tensor is available to be released.
In other words, the lifetime of a tensor is from the operation generating it to the last operation consuming it.
Let $ts$ be a tensor produced by an operation $u$, and $v_1, v_2,\dots,v_k$ be $k$ operations consuming $ts$.
The life time of $ts$ is computed as $\mathtt{max}\{\gamma(v_1), \gamma(v_2), \dots,\gamma(v_k)\} - \gamma(u)$.


%% file: graph-rewriting.tex
\begin{figure*}[t]
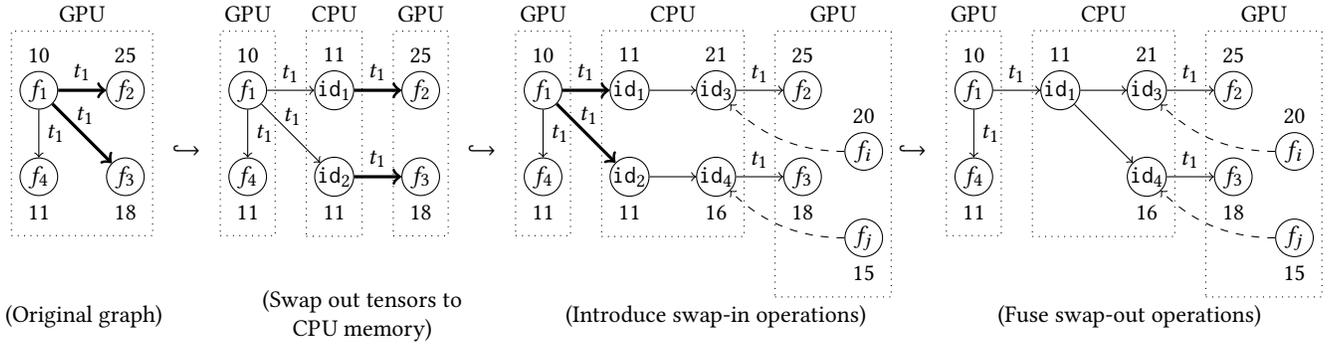

  \centering
  \tikzstyle{every node}=[scale=1, node distance=1.15cm and 1.2cm]
  \tikzstyle{ops}=[draw, circle, minimum size=0.5cm, inner sep=0]
  \tikzstyle{placeholder}=[draw, double, circle, minimum size=0.5cm, inner sep=0]
  \tikzstyle{update}=[->, dotted]
  \tikzstyle{control}=[->, dashed]
  \tikzstyle{read}=[->]
  \tikz[baseline] {
    \node at (0.6,-3) {(Original graph)};
    \node[ops, label={[name=lf1] above:10}] (f1) {$f_1$};
    \node[ops, right of=f1, label=above:25] (f2) {$f_2$};
    \node[ops, below of=f2, label=below:18] (f3) {$f_3$};
    \node[ops, below of=f1, label={[name=lf4] below:11}] (f4) {$f_4$};

    \draw[read, very thick] (f1) -- node[midway, auto] {$t_1$} (f2);
    \draw[read, very thick] (f1) -- node[midway, auto, xshift=-0.2cm] {$t_1$} (f3);
    \draw[read] (f1) -- node[midway, auto] {$t_1$} (f4);
    \node[draw, dotted,fit=(f1) (lf1) (f2) (f3) (f4) (lf4), label=above:GPU] {};
    
    \node at (4.3,-3) {\begin{tabular}{c}(Swap out tensors to \\ CPU memory) \end{tabular}};
    \node[below right of=f2] (hookarrow1) {$\hookrightarrow$};
    \node[ops, above right of=hookarrow1, label={[name=lf21] above:10}] (f21) {$f_1$};
    \node[ops, below of=f21, label={[name=lf24] below:11}] (f24) {$f_4$};
    \node[ops, right of=f21, label={[name=lid21] above:11}] (id21) {$\mathtt{id}_1$};
    \node[ops, right of=id21, label={[name=lf22] above:25}] (f22) {$f_2$};
    \node[ops, below of=id21, label={[name=lid22] below:11}] (id22) {$\mathtt{id}_2$};
    \node[ops, right of=id22, label={[name=lf23] below:18}] (f23) {$f_3$};
    \draw[read] (f21) -- node[midway, auto] {$t_1$} (id21);
    \draw[read] (f21) -- node[midway, auto, xshift=-0.2cm] {$t_1$} (id22);
    \draw[read, very thick] (id21) -- node[midway, auto] {$t_1$} (f22);
    \draw[read, very thick] (id22) -- node[midway, auto] {$t_1$} (f23);
    \draw[read] (f21) -- node[midway, auto] {$t_1$} (f24);
    \node[draw, dotted,fit=(f21) (lf21) (f24) (lf24), label=above:GPU] {};
    \node[draw, dotted, fit=(id21) (lid21) (id22) (lid22), label=above:CPU] {};    
    \node[draw, dotted,fit=(f22) (lf22) (f23) (lf23), label=above:GPU] {};

    \node at (9,-3) {(Introduce swap-in operations)};
    \node[below right of=f22] (hookarrow2) {$\hookrightarrow$};
    \node[ops, above right of=hookarrow2, label={[name=lf31] above:10}] (f31) {$f_1$};
    \node[ops, right of=f31, label={[name=lid31] above:11}] (id31) {$\mathtt{id}_1$};
    \node[ops, right of=id31, label=above:21] (id32) {$\mathtt{id}_3$};
    \node[ops, right of=id32, label={[name=lf32] above:25}] (f32) {$f_2$};
    \node[ops, below of=id32, label={[name=lid33] below:16}] (id33) {$\mathtt{id}_4$};
    \node[ops, below of=id31, label={[name=lid34] below:11}] (id34) {$\mathtt{id}_2$};
    \node[ops, right of=id33, label=below:18] (f33) {$f_3$};
    \node[ops, below of=f31, label={[name=lf34] below:11}] (f34) {$f_4$};
    \node[ops, below right of=f32, label=above:20] (f3i) {$f_i$};
    \node[ops, below right of=f33, label={[name=lf3j] below:15}] (f3j) {$f_j$};
    \node[draw, dotted,fit=(f31) (lf31) (f34) (lf34), label=above:GPU] {};
    \node[draw, dotted, fit=(id31) (lid31) (id32) (id33) (lid33), label=above:CPU] {};    
    \node[draw, dotted,fit=(f32) (lf32) (f33) (f3i) (f3j) (lf3j), label=above:GPU] {};    
    \draw[read, very thick] (f31) -- node[midway, auto] {$t_1$} (id31);
    \draw[read, very thick] (f31) -- node[midway, auto, xshift=-0.2cm] {$t_1$} (id34);    
    \draw[read] (id31) -- node[midway, auto] {} (id32);
    \draw[read] (id32) -- node[midway, auto] {$t_1$} (f32);
    \draw[read] (id34) -- node[midway, auto] {} (id33);
    \draw[read] (id33) -- node[midway, auto] {$t_1$} (f33);
    \draw[read] (f31) -- node[midway, auto] {$t_1$} (f34);
    \draw[control] (f3i) .. controls +(-1,0) and +(0.5,-0.5) .. node[midway, auto] {} (id32);
    \draw[control] (f3j) .. controls +(-1,0) and +(0.5,-0.5) .. node[midway, auto] {} (id33);

    \node at (14.5,-3) {(Fuse swap-out operations)};
    \node[right of=f3i, xshift=-0.5cm] (hookarrow) {$\hookrightarrow$};
    \node[ops, above right of=hookarrow, label={[name=lf11] above:10}] (f11) {$f_1$};
    \node[ops, right of=f11, label={[name=lid1] above:11}] (id1) {$\mathtt{id}_1$};
    \node[ops, right of=id1, label=above:21] (id2) {$\mathtt{id}_3$};
    \node[ops, right of=id2, label={[name=lf12] above:25}] (f12) {$f_2$};
    \node[ops, below of=id2, label={[name=lid3] below:16}] (id3) {$\mathtt{id}_4$};
    \node[ops, right of=id3, label=below:18] (f13) {$f_3$};
    \node[ops, below of=f11, label={[name=lf14] below:11}] (f14) {$f_4$};
    \node[ops, below right of=f12, label=above:20] (fi) {$f_i$};
    \node[ops, below right of=f13, label={[name=lfj] below:15}] (fj) {$f_j$};
    \node[draw, dotted,fit=(f11) (lf11) (f14) (lf14), label=above:GPU] {};
    \node[draw, dotted, fit=(id1) (lid1) (id2) (id3) (lid3), label=above:CPU] {};    
    \node[draw, dotted, fit=(f12) (lf12) (f13) (fi) (fj) (lfj), label=above:GPU] {};
    \draw[read] (f11) -- node[midway, auto] {$t_1$} (id1);
    \draw[read] (id1) -- node[midway, auto] {} (id2);
    \draw[read] (id2) -- node[midway, auto] {$t_1$} (f12);
    \draw[read] (id1) -- node[midway, auto] {} (id3);
    \draw[read] (id3) -- node[midway, auto] {$t_1$} (f13);
    \draw[read] (f11) -- node[midway, auto] {$t_1$} (f14);
    \draw[control] (fi) .. controls +(-1,0) and +(0.5,-0.5) .. node[midway, auto] {} (id2);
    \draw[control] (fj) .. controls +(-1,0) and +(0.5,-0.5) .. node[midway, auto] {} (id3);

  }
  \caption{Example of graph rewriting for supporting large models.
    Thick edges in left subgraph are rewritten to produce right subgraph.
    Integers above or below vertex are order of that vertex in topological ordering.
    In this example, threshold ($\alpha$) to trigger graph rewriting is $5$, so edges from $f_1$ to $f_2$ and $f_3$ are rewritten.
    $f_i$ and $f_j$ are control dependency operations that trigger executions of swap-in operations $\mathtt{id}_2$ and $\mathtt{id}_3$, respectively.
  }
  \label{fig:eg-opt}
\end{figure*}

A computational graph or a neural network model is said being large to be trained with the memory limitation of GPUs if there are many tensors that are kept in the GPU memory at a time so that they consume more memory than the GPU memory.
Hence, an out-of-memory error often happens when training such a large graph.
This is essentially because there are many tensors with a long lifetime in a computational graph.
In this section, we will show how to rewrite a large graph so that training them is possible with a limited GPU memory.
In general, our idea is temporally sending ``long lifetime'' tensors in a GPU to a CPU and sending them back to the GPU when necessary.

\subsection{Swapping  out tensors to CPU memory}
\label{sec:swap-tensor}
To put a tensor residing in GPU memory on CPU memory, we derive operations to automatically send the tensor to the CPU and send it back to the GPU.
Let us consider an edge $(f_1,f_2)$ where $\tau(f_1,f_2) = (t, \_)$ and $f_1$, $f_2$ are executed using a GPU.
Computation for this edge is
\begin{equation}
  \label{eq:1}
  f_2^{G} \circ f_1^{G}
\end{equation}
where the superscript $G$ stands for GPU.
This computation can be rewritten into:
\begin{equation}
  \label{eq:2}
  f_2^{G} \circ \mathtt{id}^{C} \circ f_1^{G}  
\end{equation}
where the superscript $C$ stands for CPU, and $\mathtt{id}$ is an identity function that is $\mathtt{id}(x) = x$.

Since $\mathtt{id}$ is executed using a CPU, the output tensor of $f_1$ will be swapped out to the CPU memory for $\mathtt{id}$ immediately after $f_1$ finishes, and GPU memory is released.
The output tensor of $\mathtt{id}$ will be swapped in to the GPU for $f_2$ when $f_2$ is triggered.
We call function $\mathtt{id}$ in Equation~\ref{eq:2} \emph{a swap-out operation}.

Using Equation~\ref{eq:2}, we are able rewrite a graph so that GPU memory consumption is reduced.
However, not all edges are needed to rewrite.
For edges $(u,v)$ where $\gamma(v) - \gamma(u) =1$, $v$ is executed immediately after $u$.
Hence, there is no need to swap the tensor on such edges.
We can define a threshold $\alpha$ and graph rewriting for an edge $u,v$ is triggered if $\gamma(v) - \gamma(u) \geq \alpha$.

\subsection{Optimization}
Equation~\ref{eq:2} is not optimized due to two reasons:
it is too late to swap the output tensor of $\mathtt{id}$ in, and $f_2$ must wait for the tensor sent from CPU memory to GPU memory;
and the tensor may be swapped out and swapped in multiple times since there may be multiple operations apart from $f_2$ reading it.
In this section we present three rules to optimize Equation~\ref{eq:2}.
Figure~\ref{fig:eg-opt} shows computational graphs obtained by each of optimization rules.

\subsubsection{Introduce swap-in operations}
To swap a tensor in early, we need an additional operation.
An Identity function can be rewritten as the composition of a function and its inverse function, that is,
\begin{equation}
  \label{eq:3}
  \mathtt{id} = f^{-1} \circ f
\end{equation}

Equation~\ref{eq:2} becomes:
\begin{equation}
  \label{eq:4}
  f_2^{G} \circ (f^{-1})^{C} \circ f^{C} \circ f_1^{G}
\end{equation}

Since $\mathtt{id}$ also has the inverse function, i.e, $\mathtt{id}$, we choose $\mathtt{id}$ for $f$ (if one would like to reduce the memory consumption on the CPU, a pair of encoding and decoding functions can be used for $f$ instead of $\mathtt{id}$),
\begin{equation}
  \label{eq:5}
  f_2^{G} \circ \mathtt{id}_2^{C} \circ \mathtt{id}_1^{C} \circ f_1^{G}
\end{equation}

In Equation~\ref{eq:5}, $\mathtt{id}_2$ will be used to swap a tensor in to a device, and we call function $\mathtt{id}_2$ \emph{a swap-in operation}.
It is worth noting that we must manually trigger $\mathtt{id}_2$ in a good order; otherwise, $\mathtt{id}_2$ is executed immediately after $\mathtt{id}_1$.
To do this, a control edge from an operation to $\mathtt{id}_2$ must be added.
We present two strategies for choosing a control operation in Section~\ref{sec:strategies}.

\subsubsection{Fuse swap-out operations}
A tensor produced by an operation is often used by multiple operations, and it is redundant if the tensor is swapped out to CPU memory multiple times.
Hence, it is recommended to always fuse swap-out operations of the same tensor into a single swap-out operation.

\subsubsection{Fuse swap-in operations}
\label{sec:fuse_swapins}
Consider a situation that multiple swap-in operations swap a tensor multiple times for multiple consuming operations.
If the tensor is large and the consuming operations are close to each other, then swapping the tensor multiple times would introduce more overhead.
In this case, it is better to fuse the swap-in operations into one swap-in operation.
The tensor is swapped in only once and resides in GPU memory to be reused by the consuming operations.
For example, in the right-most graph in Figure~\ref{fig:eg-opt}, if $f_2$ and $f_3$ are close and $t_1$ is large, then we fuse $\mathtt{id}_3$ and $\mathtt{id}_4$ into a singe swap-in operation.
To determine how close two operations are, we may define a threshold for the distance between them.

\subsection{Strategies to add control  edges}
\label{sec:strategies}
Control edges to swap-in operations are added to a computational graph to control when swap-in operations are trigger.
They are important to reduce the overhead of communication of swapping tensors in.
Consider Equation~\ref{eq:5}, a control operation for the swap-in operation $\mathtt{id}_2$ must be chosen from a set of operations, $V_c$, where $\forall v \in V_c, \gamma(\mathtt{id}_1) < \gamma(v) < \gamma(f_2)$ to guarantee the correctness of the computational graph.
Let $k = \gamma(f_2) - \gamma(v)$ be the distance between $f_2$ and $v$.
If $k$ is too small, a tensor is swapped in too late, and $f_2$ has to wait for the tensor.
If $k$ is too large, a tensor is swapped in too early, and the tensor is kept in the device for a long time before being actually used by $f_2$.

An ideal solution for choosing a control operation is having a cost model for computational graphs and using the model to prioritize operations.
However, in TensorFlow, the shape of the input and output tensors of an operation is generally unknown at the beginning unless data are fed into the graph then trigger the operation.
This means that, at the time a graph is rewritten, there is no information about the actual size of tensors, and it fails to compute operation cost statically.

In a context of statically modifying a computational graph, we introduce two parameters: \emph{lower-bound} $\sigma_l$ and \emph{upper-bound} $\sigma_u$ to handle choosing control operations.
Let us assume that an edge $(f_1, f_2)$ is rewritten using a swap-out operation $\mathtt{so}$ and swap-in operation $\mathtt{si}$:
\begin{equation}
  \label{eq:6}
  f_2^{G} \circ \mathtt{si}^{C} \circ \mathtt{so}^{C} \circ f_1^{G}
\end{equation}
We present two strategies to find a control operation for $\mathtt{si}$.

\subsubsection{Direct-order strategy}
The direct-order strategy involves directly using the topological ordering to obtain a set of candidates for control operation, starting from the target operation $f_2$ and going back to $f_1$.
Lower-bound and upper-bound are relative to $f_2$.

Algorithm~\ref{alg:directorder} shows the algorithm of this strategy.
Candidates are operations whose distance to $f_2$ is in the range of $\sigma_l$ to $\sigma_u$ (Line $7$) and there exists a path from them to $f_2$ (Line $8$).
The algorithm stops once it has found one operation satisfying the above conditions (Lines $9$--$12$).

\begin{algorithm}[t]
  \centering
  \caption{Direct-order strategy}
  \label{alg:directorder}
  \renewcommand{\algorithmicrequire}{\textbf{Input:}}
  \renewcommand{\algorithmicensure}{\textbf{Output:}}
  \begin{algorithmic}[1]
    \Require{source operation $f_1$, target operation $f_2$, lower-bound $\sigma_l$, upper-bound $\sigma_u$}
    \Ensure{an operation $g$}
    \State $l \gets \mathtt{max}\{\gamma(f_2) - \sigma_u + 1, \gamma(f_1)\}$ \Comment{Lowest order}
    \For{$i \gets \sigma_l$ to $\sigma_u$}
    \State $k \gets \gamma(f_2) - i$ \Comment{$i$ operations before $f_2$}
    \If{$k \leq l$} \Comment{Out of range}
    \State \Return \text{null}
    \EndIf
    \State $T \gets \{f ~|~ f \in V, \gamma(f) = k\}$ \Comment{All operations of order $i$}
    \State $T \gets \{f ~|~ f \in T, f_2 \in \mathtt{Reach}(f)\}$ \Comment{$f_2$ is reachable from $f$}
    \If{$T$ is not empty}
    \State $g \gets \Call{GET}{T}$ \Comment{Randomly get one item in $T$}
    \State\Return $g$
    \EndIf
    \EndFor
  \end{algorithmic}
\end{algorithm}

\subsubsection{Chain-rule strategy}
The chain-rule strategy involves starting from the source operation $f_1$ and going down along the forward phase to find corresponding backward operations as candidates for control operations.
Breadth-first search is used to traverse operations in the forward phase in which lower-bound and upper-bound are used to limit the search space of forward operations.
In other words, lower-bound and upper-bound are relative to the source operation $f_1$.

Algorithm~\ref{alg:chainrule} shows the algorithm of this strategy.
For breadth-first search, we maintain two open sets $S_1$ and $S_2$, and one closed set $S_c$.
The $S_1$ contains current forward operations, and
$S_2$ contains forward operations for the next level (including all outgoing operations of operation in $S_1$).
The $S_c$ contains visited operations.
Starting from $f_1$, once the algorithm is in the range of $\sigma_l$ to $\sigma_u$ (Line $8$), it obtains outgoing backward operations of a current operation (Line $9$), then checks the validity of these backward operations (Lines $10$--$11$). If there is one valid operation, it is a candidate and the algorithm returns it.
Otherwise, the algorithm goes to the next level (Lines $27$--$30$).

\begin{algorithm}[t]
  \centering
  \caption{Chain-rule strategy}
  \label{alg:chainrule}
  \renewcommand{\algorithmicrequire}{\textbf{Input:}}
  \renewcommand{\algorithmicensure}{\textbf{Output:}}
  \begin{algorithmic}[1]
    \Require{source operation $f_1$, target operation $f_2$, lower-bound $\sigma_l$, upper-bound $\sigma_u$}
    \Ensure{an operation $g$}
    \State $S_1 \gets \{f_1\}$; $S_2 \gets \emptyset$; $S_c \gets \emptyset$
    \While{$S_1$ is not empty}
    \If{$\sigma_u = 0$ \textbf{or} $\sigma_l > \sigma_u$}
    \State \Return \text{null}
    \EndIf 

    \State $s \gets$ \Call{GET}{$S_1$} \Comment{Get one item in $S_1$}
    \State $T \gets$ \Call{Out}{$s$} \Comment{Outgoing operations of $s$}

    \If{$\sigma_l \leq 0$} \Comment{Inside the range}
    \State $B \gets \{f ~|~ f \in T, f \text{ is a backward operation}\}$
    \State $B \gets \{f ~|~ f \in B, \gamma(f) > \gamma(f_1), \gamma(f) < \gamma(f_2)\}$
    \State $B \gets \{f ~|~ f \in B, f_2 \in \mathtt{Out}(f)\}$ \Comment{$f_2$ is reachable from $f$}
    \If{$B$ is not empty}
    \State $g \gets$ \Call{GET}{B} \Comment{Randomly get one item in $B$}
    \State \Return $g$
    \EndIf
    \EndIf
    
    \State $N \gets \{f ~|~ f \in T, f \text{ is a forward operation}\}$

    \For{$f$ in $N$}
    \If{$f \in S_c$} \Comment{$f$ is visited}
    \State \textbf{continue}
    \EndIf
    \If{$f \notin S_2$}
    \State $S_2 \gets S_2 \cup \{f\}$
    \EndIf
    \EndFor

    \State $S_c \gets S_c \cup \{s\}$ \Comment{mark $s$ as visited}

    \If{$S_1$ is empty} \Comment{go down one level}
    \State $\sigma_l \gets \sigma_l - 1$; $\sigma_u \gets \sigma_u - 1$
    \State $S_1 \gets S_2$; $S_2 \gets \emptyset$
    \EndIf
    \EndWhile
  \end{algorithmic}
\end{algorithm}


%% file: tflms.tex
\begin{figure}[!ht]
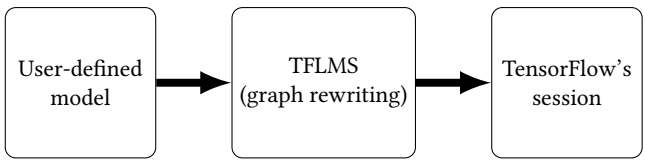

  \centering
  \tikzstyle{mybox}=[draw, rectangle, minimum size=2cm, rounded corners]
  \tikzstyle{myarrow}=[->, >=latex, line width=3pt]
  \tikz[every text node part/.style={align=center}]{
    \node[mybox] (n1) {User-defined \\ model};
    \node[mybox, right=of n1] (n2) {TFLMS \\ (graph rewriting)};
    \node[mybox, right=of n2] (n3) {TensorFlow's \\ session};

    \draw[myarrow] (n1) -- (n2);
    \draw[myarrow] (n2) -- (n3);
  }
  \caption{TFLMS module in TensorFlow.}
  \label{fig:tflms-arch}
\end{figure}

\begin{table*}[!ht]
  \centering
  \caption{Parameters in TFLMS}
  \tikzstyle{every node}=[scale=1]
  \tikzstyle{ops}=[draw, circle, minimum size=0.5cm, inner sep=0]
  \tikzstyle{placeholder}=[draw, double, circle, minimum size=0.5cm, inner sep=0]
  \begin{tabularx}{\linewidth}{lLc}
    \toprule
    Parameter & Meaning & Default value\\
    \midrule
    graph & The graph we will modify for LMS. This should be the graph of user-defined neural network.& required \\ \midrule

    optimizer\_scopes & A set of scopes for the optimizers/solvers. & required\\ \midrule

    starting\_scope & Tensors that are reachable from the operations in this scope will be swapped for LMS. Set this to the scope of the first layer if we would like to modify the whole graph. & None \\ \midrule

    starting\_op\_names & Tensors that are reachable from the operations with these names will be swapped for LMS. & None \\ \midrule

    excl\_scopes & A set of scopes. Output tensors of operations in the scopes will not be swapped out to CPU memory. & empty\\ \midrule

    incl\_scopes & A set of scopes. Output tensors of operations in the scopes will be swapped out to CPU memory. & empty\\ \midrule

    excl\_types & A set of types. Output tensors of operations with these types will not be swapped out to CPU memory. & empty\\ \midrule

    incl\_types & A set of types. Output tensors of operations with these types will be swapped out to CPU memory. & empty\\ \midrule

    n\_tensors & The number of tensors for LMS, counting from the starting\_scope. & -1 (all tensors)\\ \midrule

    lb & Lower-bound value for LMS. & 1\\ \midrule

    ub & Upper-bound value for LMS. & 10000\\ \midrule

    ctrld\_strategy & Two strategies to find control dependency operations for swap-in operations: chain\_rule and direct\_order. & chain\_rule\\ \midrule

    fuse\_swapins & Fuse "close" swap-in operations into one operation. & False\\ \midrule
    
    swap\_branches & If True, LMS will swap tensors in branches in the forward phase. & False\\ \midrule

    branch\_threshold & A threshold for swapping branches in the forward phase. & 0\\
    \bottomrule
  \end{tabularx}
  \label{tab:tflms:params}
\end{table*}

We developed a TensorFlow module, named \emph{TFLMS}, based on our proposed approach.
The module allows users to quickly turn their large model into one that can be trained with limited GPU memory.
In TensorFlow, users first define a neural network model.
TensorFlow then automatically generates a computational graph from the model.
Finally, users define a TensorFlow session to execute operations in the computational graph.
Once a session is invoked, users cannot modify the computational graph.
Hence, we implement TFLMS to statically modify the graph before a session starts.

Figure~\ref{fig:tflms-arch} shows how TFLMS is positioned in TensorFlow.
TFLMS takes a computational graph and automatically modifies it using the transformation rules presented in Section~\ref{sec:grprw}.
TFLMS uses APIs in the module ``graph editor''\footnote{Graph editor: \url{https://www.tensorflow.org/api_guides/python/contrib.graph_editor}} in TensorFlow to modify the graph.
The modified graph is then executed by a TensorFlow session as normal.
TFLMS's source code is publicly available as a pull request in the TensorFlow repository\footnote{https://github.com/tensorflow/tensorflow/pull/19845}.

\begin{lstlisting}[language=Python, numbers=left, stepnumber=1, caption={Sample Python code to use TFLMS in TensorFlow.}, label={lst:tflms}, float=tp]
# define a scope for the optimizer/solver
with tf.name_scope('adam_optimizer'):
   opt = tf.train.AdamOptimizer(1e-4)
   train_step = opt.minimize(cross_entropy)

# define a LMS instance and run it
from tensorflow.contrib.lms import LMS
lms_obj = LMS({'adam_optimizer'})
lms_obj.run(graph=tf.get_default_graph())

with tf.Session() as sess:
    sess.run(tf.global_variables_initializer())
	batch = mnist.train.next_batch(50)
	train_step.run(feed_dict={x: batch[0],
                                  y_: batch[1]})
\end{lstlisting}

Listing~\ref{lst:tflms} shows a brief example of using TFLMS in TensorFlow.
While defining a neural network, users must define a scope for their optimizer (Line $2$).
Users then define a LMS instance for that scope and run the instance to modify the computational graph of the neural network (Lines $7$--$9$).
After that, users create a TensorFlow session and train the network as usual.

\subsection{Implementation}
The important part of TFLMS is building a topological ordering.
Given a graph, TFLMS uses the python package ``toposort''\footnote{https://pypi.org/project/toposort/} to build a topological order.
The topological ordering, $\gamma$, is to decide which tensors are swapped out and when they are swapped in as shown Section~\ref{sec:grprw}.
To rewrite edges, TFLMS traverses through the graph using the breadth-first search algorithm, starting from input variables.
We do not rewrite incoming and outgoing edges of variables.
In other words, learnable parameters are kept in GPU memory.
Apart from an input of a computational graph, TFLMS allows users to pass other parameters to flexibly control how the graph is modified.
Table~\ref{tab:tflms:params} lists the parameters in TFLMS.

By default, TFLMS always rewrites edges between a forward operation and a backward operation.
To determine operations in the backward phase, users should pass the scope\footnote{In TensorFlow, scope defines a name for a set of operations, similar to a folder in a file system.} of solvers or optimizers that are used to train the model (via TFLMS parameter \emph{optimizer\_scopes}).
Note that, it is possible to automatically rewrite the whole graph without \emph{optimizer\_scopes}.
Using \emph{optimizer\_scopes} reduces unnecessary operations that are not helpful for large model support, e.g. operations in the update phase.
If a model has many branches in the forward phase, users may want to use parameters \emph{swap\_branches} and \emph{branch\_threshold} to enable rewriting edges $(u,v)$ satisfying $\gamma(v) - \gamma(u) > \text{\emph{branch\_threshold}}$.
\emph{branch\_threshold} is the threshold $\alpha$ defined in Section~\ref{sec:swap-tensor}.
Swapping tensors in the forward phase may affect the performance of inferencing of a neural network because it introduces overhead of swapping the tensors out and in.
However, if the neural network is still large for inferencing, swapping those tensors is necessary.
Without enabling \emph{swap\_branches}, our modification does not cause any affect on the performance of inferencing because added swap-out and swap-in operations between the forward and backward phases are not executed during the inferencing.
Inclusion or exclusion of an operation can be done via the operation's type or scope.
Users can define a starting point for the breadth-first search by using the scope or name of operations via parameters \emph{starting\_scope} and \emph{starting\_op\_names}.
By default, TFLMS rewrites all reachable edges.
However, users can define the number of tensors that are swapped via parameter \emph{n\_tensor}.
Parameters \emph{lb} and \emph{ub} are lower-bound and upper-bound, respectively, as defined in Section~\ref{sec:strategies}.
A strategy for choosing control operations is set by parameter \emph{ctrld\_strategy}.
Parameter \emph{fuse\_swapins} is to enable the optimization of fusing swap-in operations.

\subsection{Performance tuning}

To get the maximum performance when using TFLMS, we need to find the combination of tuning parameters that provides the fastest training time with the model.
The goal of the performance tuning is to swap out enough tensors to allow our training to run without out-of-memory errors, while not swapping too many such that the extra swapping communication overhead degrades performance.

The two tuning parameters we should focus on are \emph{n\_tensors} and \emph{lb}.
Since \emph{n\_tensors} controls the number of tensors that will be swapped, the higher this is set, the lower the peak GPU memory usage will be.
The \emph{lb} controls how soon the tensor is swapped back in before use.
A low value \emph{lb} can make the training on the GPU pause and wait while the swap-in finishes.
This will degrade performance.
A higher value \emph{lb} allows the tensor swap-in to finish before it is needed and allows training to run without pause.
The downside to swapping in too early is that more tensors will be in the GPU memory at any point in time, resulting in higher peak GPU memory usage.

Tuning thus becomes finding the correct balance between \emph{n\_tensors} and \emph{lb} that provides the best performance for a given model.
To start the performance tuning it is suggested that \emph{n\_tensors} be set to -1, which will swap all reachable tensors, e.g., $N$ tensors.
The \emph{lb} should be set to the default of 1, which is the latest possible swap-in.
It is useful to run with $\emph{n\_tensors}=N$ and then adjust it downward.
If the model has branches similar to the 3UNet model, it is likely useful to set \emph{swap\_branches} to True and tune the branch threshold.


%% file: experiments.tex
\subsection{Experimental environment}
Experiments were run on an IBM POWER8 NUMA-based machine~\cite{s82lc-hpc} using one GPU.
%
%
The machine has two 4GHz 10-core POWER8 processors, eight simultaneous multi-threads (SMTs) per core and 256 MB RAM per processor.
There are four NVIDIA Tesla P100 GPUs (each with 16 GB memory).
NVLinks are used for connections among GPUs and CPUs: one 80 GB/s duplex link between GPUs 0 and 1, one 80 GB/s duplex link between GPUs 2 and 3, two 80 GB/s duplex links from CPU 0 to GPUs 0 and 1, and two 80 GB/s duplex links from CPU 1 to GPUs 2 and 3.
On the machine, we installed TensorFlow 1.8, CUDA Toolkit v9.0 and cuDNN 7.0.5.

We evaluated TFLMS using two popular neural networks: ResNet-50 for image recognition and 3DUNet for image segmentation.
To make a model larger, we increase the batch size of each iteration.
By default, we always fuse swap-out operations.

\subsection{Maximum batch size}
\begin{table}[!t]
  \centering
  \caption{Maximum batch size when swapping all reachable tensors. OOM stands for out-of-memory.}  
  \begin{tabular}{ccccc}
    \toprule
    Model & Image & Without TFLMS & With TFLMS & Ratio\\
    \midrule
    ResNet-50 & $224^2$ & $176$ & $832$ & $4.7$\\
    3DUnet & $128^3$ & $2$ & $4$ & $2$\\
    3DUnet & $192^3$ & OOM & $1$ & $$\\
    \bottomrule
  \end{tabular}
  \label{tab:maxbs}
\end{table}

Table~\ref{tab:maxbs} shows the maximum batch size we are able to train using TFLMS.
We let TFLMS swap all reachable tensors to reduce GPU memory consumption as much as possible.
In total, TFLMS swapped all of $317$ tensors for ResNet-50, all of $2779$ tensors for 3DUNet with $192^3$ images and $2397$ tensors for 3DUNet with $128^3$ images\footnote{3DUnet architecture is changed according as image size.}.
With TFLMS we were able to train ResNet-50 and 3DUNet with $4.7$ and $2$ times larger batch size, respectively.
For 3DUnet, we were able to train the whole images of $192^3$ without resizing or splitting the images, which was impossible without TFLMS.

\subsection{Training performance}
\begin{figure}[!ht]
  \centering
  \includegraphics[scale=0.3,angle=-90]{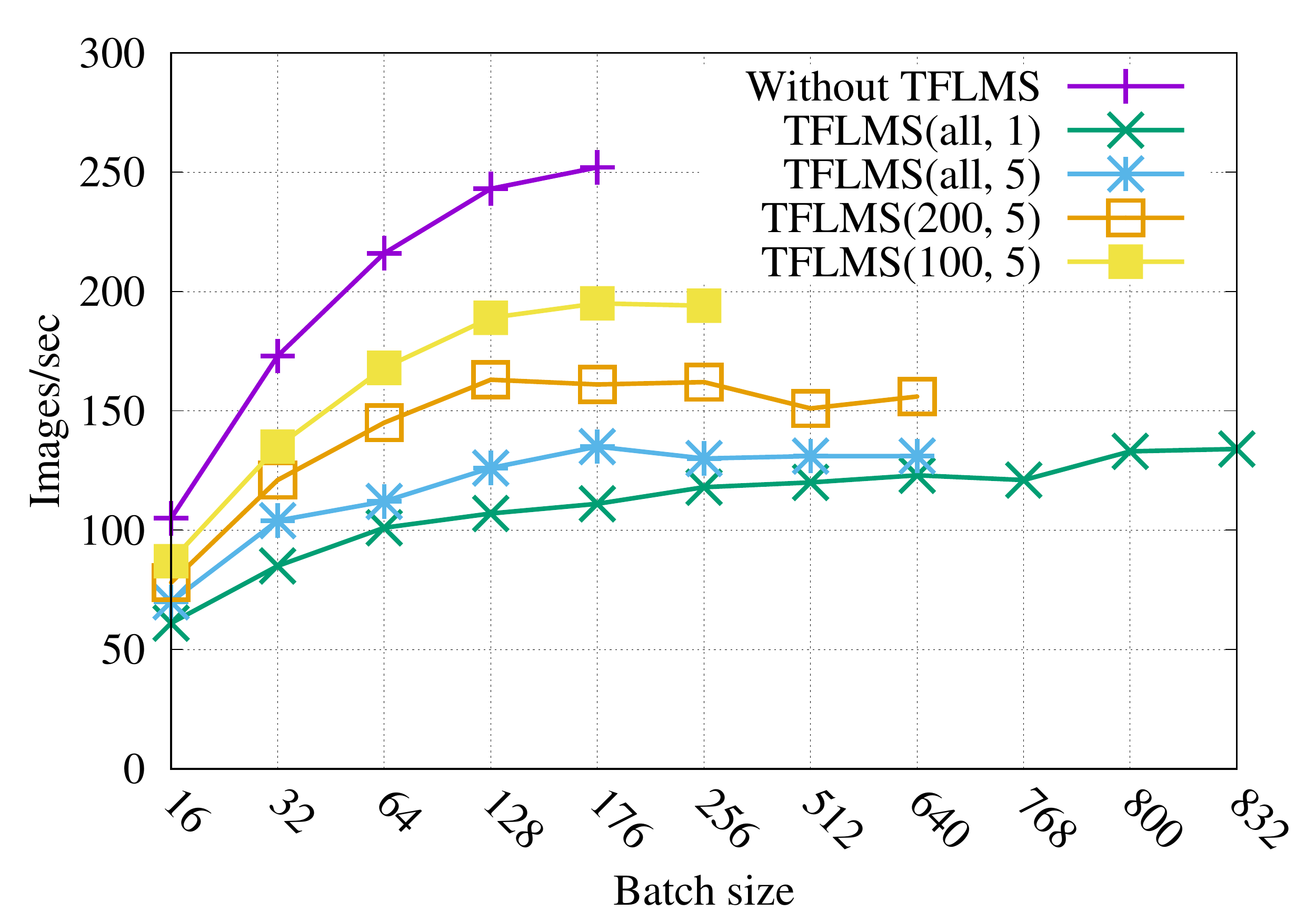}
  \caption{Effectiveness of n\_tensors and lb on training performance of ResNet-50. TFLMS(x, y) means running TFLMS with n\_tensors=x and lb=y. ``all'' means swapping all tensors, in this case n\_tensors=317.}
  \label{fig:resnet-perf}
\end{figure}

\begin{figure}[!ht]
  \centering
  \includegraphics[scale=0.3,angle=-90]{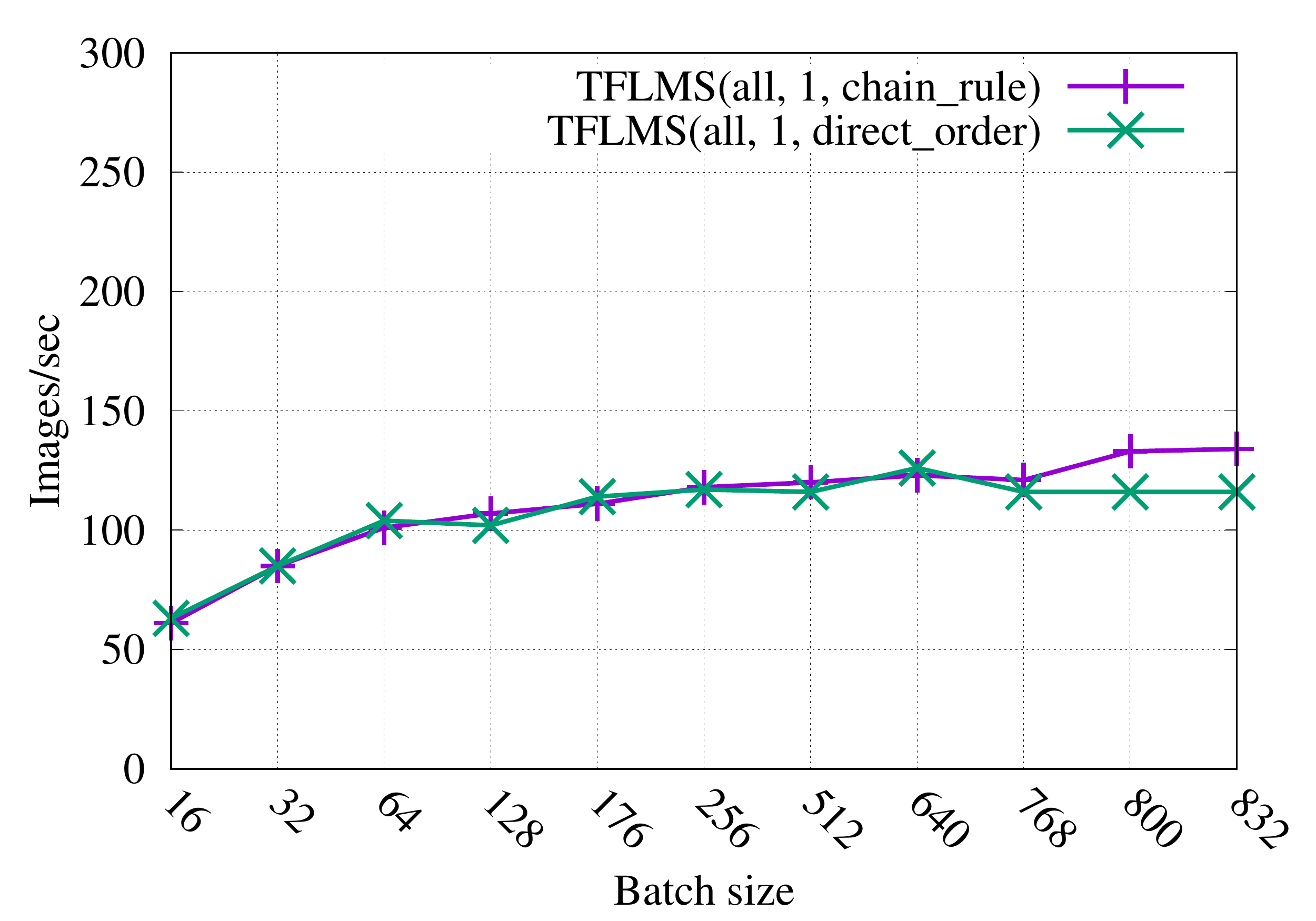}
  \caption{Effectiveness of ctrld\_strategy on training performance of ResNet-50. TFLMS(x, y, z) means running TFLMS with n\_tensors=x, lb=y, ctrld\_strategy=z. ``all'' means swapping all tensors, in this case n\_tensors=317.}
  \label{fig:resnet-strategy}
\end{figure}

\begin{figure}[!ht]
  \centering
  \includegraphics[scale=0.3,angle=-90]{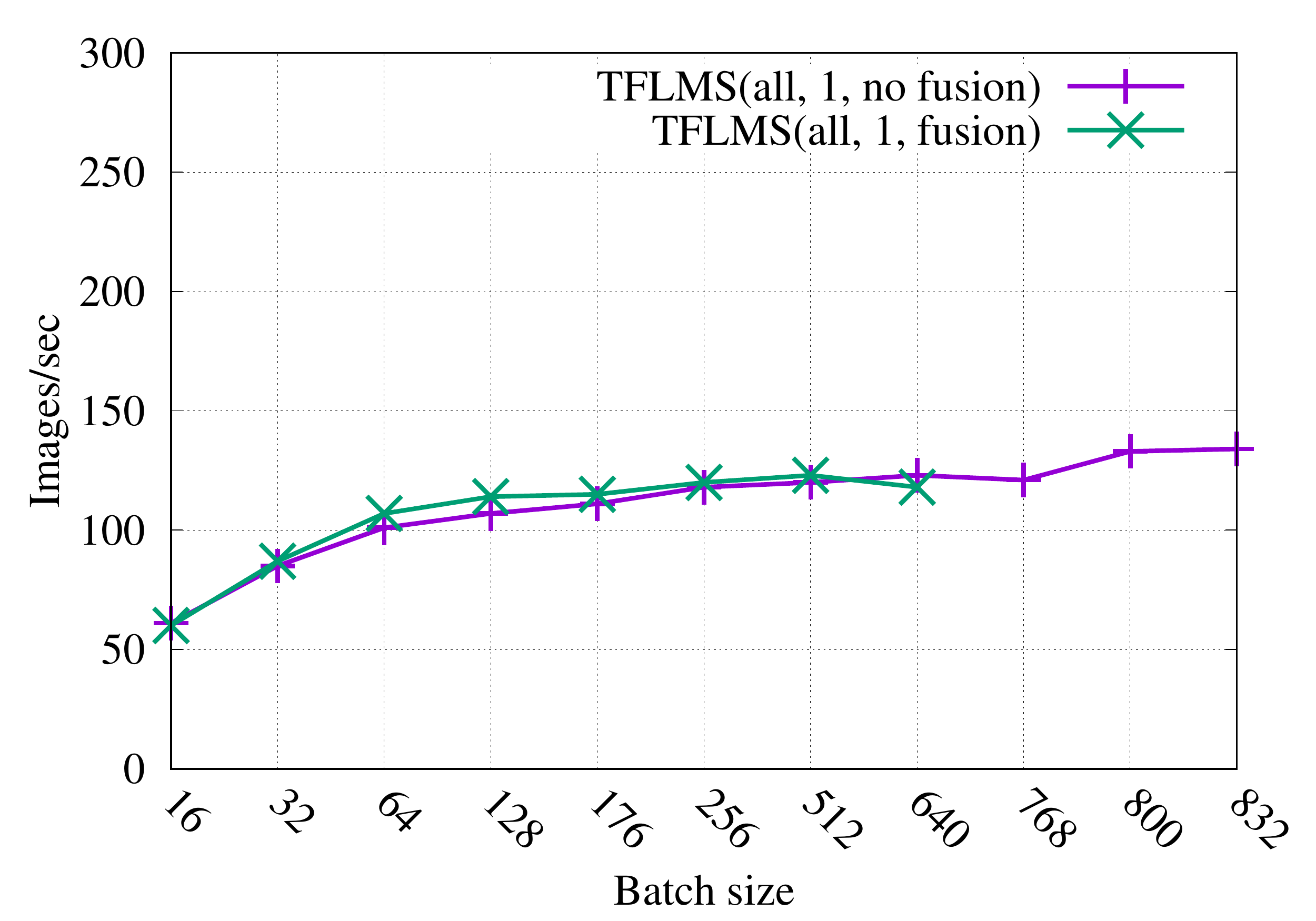}
  \caption{Effectiveness of fuse\_swapins on training performance of ResNet-50. TFLMS(x, y, z) means running TFLMS with n\_tensors=x, lb=y, fused\_swapins=z. ``all'' means swapping all tensors, in this case n\_tensors=317.}
  \label{fig:resnet-fusion}
\end{figure}

\begin{figure}[!ht]
  \centering
  \includegraphics[scale=0.3,angle=-90]{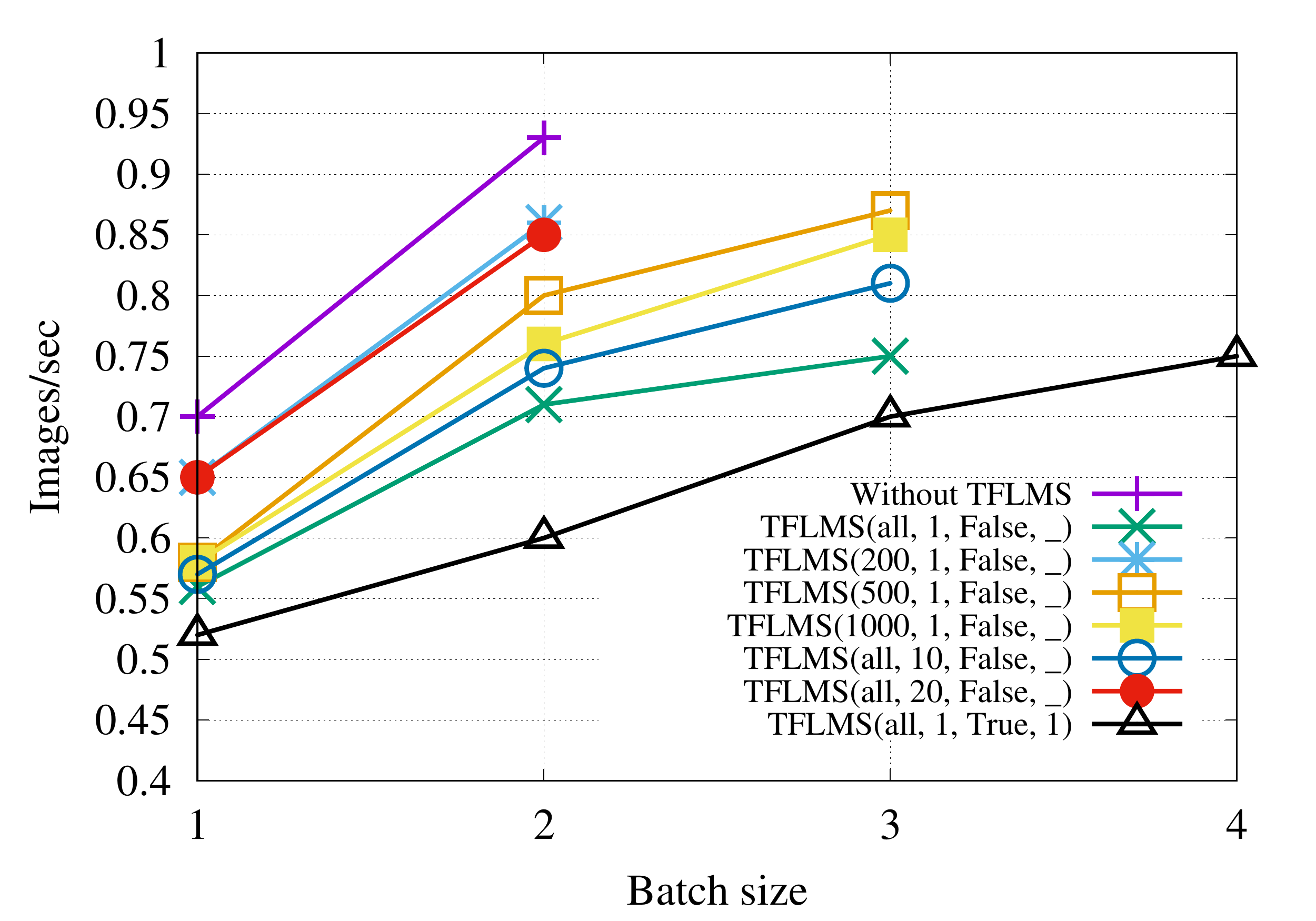}
  \caption{Effectiveness of n\_tensors, lb, swap\_branches on training performance of 3DUnet. TFLMS(w, x, y, z) means running TFLMS with n\_tensors=w, lb=x, swap\_branches=y, branch\_threshold=z. ``all'' means swapping all tensors, in this case n\_tensors=$2397$. Input images are of size of $128^3$.}
  \label{fig:3dunet-128-perf}
\end{figure}

Figure~\ref{fig:resnet-perf} shows the effectiveness of parameters n\_tensors and lb on training  performance of ResNet-50.
We measured the number of images per second (images/sec) for each batch size.
Without TFLMS, the maximum batch size we were able to train is $176$.
Performance for a smaller batch size was poor because GPU usage was small.
With TFLMS, when we first swapped out all reachable tensors, i.e. $317$ tensors, and set \textit{lb} to $1$ for swapping in a tensor as late as possible, the maximum batch size we were able to train is $832$, $4.7$ times larger than the one without TFLMS.
However, performance was not good.
We then tried to increase \textit{lb} from $1$ to $5$ to swap in tensor earlier so that there were more overlap between computation and communication.
It is clear that the higher \textit{lb}, the better training performance, but the maximum batch size was decreased because there were more tensors residing in GPU memory at a time.
Similarly, we decreased the number of tensors being swapped out, from $317$ (all) to $200$ or $100$.
We also obtained better performance.
\textit{n\_tensors} was more effective than \textit{lb} on training performance, and \textit{lb} was less effective than \textit{n\_tensors} on the maximum batch size.
Hence, there should be a tradeoff between \textit{n\_tensors} and \textit{lb}.

Figure~\ref{fig:resnet-fusion} shows the effectiveness of fusing swap-in operations.
In both cases, we swapped out $317$ tensors in total, but the numbers of swapping operations added to the graph with \emph{fuse\_swapins} enabled and disabled are $687$ and $634$, respectively.
Fusing swap-in operations lead to better performance but smaller maximum batch size.
This is because some tensors were kept in GPU memory for re-using as we mentioned in Section~\ref{sec:fuse_swapins}.

Figure~\ref{fig:resnet-strategy} shows a comparison between two strategies ``chain\_rule'' and ``direct\_order'' for finding control dependency operations.
Though the strategy ``direct\_order'' is simple than ``chain\_rule'', it sometimes had poorer performance for training ResNet-50.
In particular, ``direct\_order'' was much slow with batch sizes $768$, $800$ and $832$.

Figure~\ref{fig:3dunet-128-perf} shows results for 3DUnet.
The maximum batch size we were able to train with TFLMS is twice as large as that without TFLMS.
The effectiveness of Parameters n\_tensors and lb for 3DUnet is similar to that for ResNet-50.
In particular, when we decreased n\_tensors from $2397$ (all tensors) to $1000$, we clearly saw better performance, but the maximum batch size was decreased from $4$ to $3$.
We measured the effectiveness of swapping branches.
We enabled swapping branches with threshold $20$, the number of added operations was increased from $3895$ to $4052$ and the number of swapped tensors stayed the same.
By swapping branches, we were able to train 3DUnet with the maximum batch size of $4$ instead of $3$.
We also tried to train 3DUnet with large images, i.e. images of size of $192^3$.
While without TFLMS we got out-of-memory errors, with TFLMS, we were able to train 3DUnet at $0.17$ images/sec (Batch size=$1$, n\_tensors=$2779$ (all), lb=$1$, swap\_branches=True, branch\_threshold = $20$).


%% file: conclusion.tex
We have proposed a formal approach to deriving swap-out and swap-in operations for enabling large model support.
We formally revised the concept of computational graph and borrowed the theory of program transformations to derive new operations as well as optimize the graph.
Furthermore, We have proposed two strategies to statically find control dependency operations for triggering swap-in operations.
The experimental results showed that our approach helped train very large models, i.e. $4.7$ and $2$ times larger for ResNet-50 and 3DUnet, respectively.
Though our definition of computational graph is inspired by TensorFlow, it is still general enough to be applied to other computational graph based frameworks.
In the future, we plan to incorporate the re-computation technique by introducing new transformation rules.
Investigating a good heuristics to finding control dependency operations is an open problem.

%% file: acks.tex
Authors would like to thank Samuel D. Matzek from IBM Systems PowerAI team for helping re-factor our source code for the pull request.
The authors would also like to thank Geert Janssen and Minsik Cho from IBM Research for their fruitful discussion on our approach for large model support.
